# FairMarket-RL: LLM-Guided Fairness Shaping for Multi-Agent Reinforcement Learning in Peer-to-Peer Markets

*Shrenik Jadhav[1], Birva Sevak[1], Srijita Das[1], Akhtar Hussain[2], Wencong Su[3,\*], Van-Hai Bui[3,\*]*

[1]*Departmernt of Computer and Information Science, University of Michigan-Dearborn, USA*

[2]*Departmernt of Electrical and Computer Engineering, Laval University, Canada*

[3]*Departmernt of Electrical and Computer Engineering, University of Michigan-Dearborn, USA*

**Abstract:** Peer-to-peer (P2P) trading is increasingly recognized as a key mechanism for decentralized market regulation, yet existing approaches often lack robust frameworks to ensure fairness. This paper presents FairMarket-RL, a novel hybrid framework that combines Large Language Models (LLMs) with Reinforcement Learning (RL) to enable fairness-aware trading agents. In a simulated P2P microgrid with multiple sellers and buyers, the LLM acts as a real-time fairness critic, evaluating each trading episode using two metrics: Fairness-To-Buyer (FTB) and Fairness-Between-Sellers (FBS). These fairness scores are integrated into agent rewards through scheduled λ-coefficients, forming an adaptive LLM-guided reward shaping loop that replaces brittle, rule-based fairness constraints. Agents are trained using Independent Proximal Policy Optimization (IPPO) and achieve equitable outcomes, fulfilling over 90% of buyer demand, maintaining fair seller margins, and consistently reaching FTB and FBS scores above 0.80. The training process demonstrates that fairness feedback improves convergence, reduces buyer shortfalls, and narrows profit disparities between sellers. With its language-based critic, the framework scales naturally, and its extension to a large power distribution system with household prosumers illustrates its practical applicability. FairMarket-RL thus offers a scalable, equity-driven solution for autonomous trading in decentralized energy systems.

## 1.Introduction

Peer-to-peer (P2P) markets allow distributed participants to trade directly with one another, without any central clearing authority [1],[2]. Decentralisation boosts resilience and autonomy, but it also surfaces fairness concerns: some sellers can dominate profits, or buyers may pay systematically higher prices. Recent multi-agent reinforcement-learning (MARL) prototypes demonstrate that decentralised bidding and clearing are technically feasible, yet sizeable payoff disparities persist; for example, Chen and Liu's networked-MARL platform needed ad-hoc penalties to keep offer prices within socially acceptable bounds [3].

Traditional attempts to mitigate such inequity embed fairness directly into the reward with static, hand-crafted rules. Potential-based shaping terms, proportional-fair scoring, or profit-variance penalties can tame extreme outcomes in controlled testbeds [4],[5],[6], but they are brittle and difficult to port as market conditions evolve. Subsequent work replaced these heuristic tweaks with more principled objectives: Siddique *et al.* recast the problem as multi-objective RL and optimised a lexicographic-maximin welfare; Zimmer *et al.* introduced a decentralised equity network that learns an additional shaping term alongside the policy. Although both studies report roughly 20 % improvements in fairness metrics, they still require painstaking metric design and sensitive hyper-parameter tuning [7],[8].

Meanwhile, large language models (LLMs) offer a qualitatively different avenue. Instruction-tuned systems such as InstructGPT absorb broad moral and economic priors from human-feedback data and can reason about equity in free-form language [9]. Although LLMs fine-tuned with human feedback already align well with social objectives in text-generation tasks, using an LLM as a live moral critic in non-text MARL loops remains largely unexplored; most fairness-aware PPO variants still rely on explicit penalty terms or demographic-parity constraints, underscoring the need for a more general, low-friction feedback source. On the RL side, Proximal Policy Optimisation (PPO) and its independent variant IPPO, which assigns each agent its own critic has become a de-facto baseline for decentralised control because it combines stable clipped-surrogate updates with minimal coordination assumptions [10]. IPPO's on-policy nature meshes naturally with per-episode reward modulation, making it an ideal backbone for fairness-shaped MARL.

However, algorithmic stability alone does not guarantee equity [11], [12]. Left to optimise purely monetary rewards, PPO/IPPO agents still gravitate toward profit maximising strategies that widen payoff gaps and erode buyer welfare. What is missing is a plug-and-play fairness critic capable of evaluating each trading episode through a human-centric lens and issuing dense, differentiable feedback that learning updates can exploit. Recent progress in large language models offers such a solution: an LLM can digest a structured summary of market outcomes and instantly return nuanced assessments of distributive justice, without the need for hand-crafted metrics or costly human annotation [13].

Building on IPPO's stable on-policy updates, we present FairMarket-RL, a fully decentralised trading framework that injects a state-of-the-art instruction-tuned LLM into the reward loop as a real-time fairness critic. After each episode the model ingests a compact summary of prices, quantities, profits, margins, unsold inventory, and unmet demand, then emits two scalar signals—Fairness-to-Buyer (FTB) and Fairness-Between-Sellers (FBS) that are blended into every agent's reward through scheduled λ-coefficients. These signals simultaneously encourage high demand satisfaction, balanced seller margins, profit parity, and anti-monopoly behaviour by penalising lopsided sales shares or excessive mark-ups. Extensive simulations show that this purely LLM-driven shaping pushes demand fulfilment above 90 %, keeps both fairness scores about 0.80, maintains healthy profit margins for each seller, and prevents single-seller dominance without hand-crafted rules or human oversight. FairMarket-RL is the first system to embed an LLM's moral reasoning directly into multi-agent reward shaping, charting a scalable path toward resilient, socially aligned P2P markets.

## 2. Problem Statement

We generalise FairMarket-RL to a market with $N_S$ sellers and $N_B$ buyers. The environment is modelled as a finite-horizon, turn-based game

$$G = \left\langle \mathcal{S}, \{\mathcal{A}^{\{S_i\}}\}_{\{i=1\}}^{\{N_S\}}, \{\mathcal{A}^{\{B_j\}}\}_{\{j=1\}}^{\{N_B\}}, T, R \right\rangle \quad (1)$$

where $\mathcal{S}$ is the global state space; $\mathcal{A}^{\{S_i\}}$ and $\mathcal{A}^{\{B_j\}}$ are the action spaces of Seller $i$ and Buyer $j$; $T$ is a deterministic transition function; and $R$ returns the terminal reward vector for all agents. Each episode, a single trading round unfolds in stages: sellers act in order $(S^1 \ldots S_{\{N_S\}})$, posting price quantity offers without seeing later sellers' decisions; buyers then act sequentially, allocating their demand over the posted offers. This structure captures fairness and efficiency issues in peer-to-peer markets with many participants.

### 2.1 Environment dynamics

- State space $\mathcal{S}$. At timestep $t$ the global state

$$s_t = [I^1(t), \ldots, I_{\{N_S\}(t)}, D^1(t), \ldots, D_{\{N_B\}(t)}, p^1(t), q^1(t), \ldots, p_{\{N_S\}(t)}, q_{\{N_S\}(t)}, \sigma_t] \quad (2)$$

where $I_{i(t)}$ is Seller $i$'s remaining inventory, $D_{j(t)}$ is Buyer $j$'s residual demand, $(p_{i(t)}, q_{i(t)})$ is Seller $i$'s current offer (initially zero), and $\sigma_t \in \{1, \ldots, N_S + N_B\}$ labels the stage $(1 \ldots N_S = sellers, N_S + 1 \ldots = buyers)$. A seller observes its own inventory and all buyers' demands but not previous sellers' offers; a buyer observes the full state when its turn arrives.

- Action spaces. *Seller i* chooses a price quantity offer $(p_i, q_i) \in \{1, \ldots, 10\} \times \{0, \ldots, I_{i(t)}\}$, ensuring $q_i \leq I_i$. *Buyer j* chooses a non-negative allocation $vector\ b^{\{(j)\}} = \left(b^{\{(j)\}1}, \ldots, b^{\{(j)\}}_{\{N_S\}}\right)$ with $\Sigma_i b_i^{\{(j)\}} \leq D_{j(t)}$ and element wise constraint $b_i^{\{(j)\}} \leq q_i$, buyers cannot purchase more than the units offered by a seller.

- Transition $T$. After all buyer's act, inventories and residual demands update deterministically:

$$I_{i(t+1)} = I_{i(t)} - \Sigma_j b_i^{\{(j)\}}, \quad D_{j(t+1)} = D_{j(t)} - \Sigma_i b_i^{\{(j)\}} \tag{3}$$

## 2.2 Raw reward formulation

At episode termination, we first compute each agent's raw economic reward, before any fairness shaping is applied:

$$r_{\{S_i\}}^{\{raw\}} = (p_i - c) \cdot \Sigma_j b_i^{\{(j)\}} - \alpha \cdot D_{\{unsat\}} - \beta \cdot I_{\{i, unsold\}} \tag{4}$$

where $c$ is unit cost, $D_{\{unsat\}} = \Sigma_j D_{j(t_{end})}$ is total unmet demand, and $I_{\{i, unsold\}} = I_i - \Sigma_j b_i^{\{(j)\}}$ is *Seller i's* leftover inventory.

For each buyer $Bj$, the raw payoff combines total spending with the same demand-shortfall penalty:

$$r_{\{B_j\}}^{\{raw\}} = -\Sigma_i p_i b_i^{\{(j)\}} - \alpha \cdot D_{\{unsat\}} \tag{5}$$

The shared penalty α makes all agents jointly responsible for meeting demand.

## 2.3 LLM-based fairness shaping

To incorporate fairness considerations that are not captured by pure profit and penalty terms, we augment the raw rewards with guidance from a language-model critic f_{LLM}. After every episode the model returns two scalar scores—FTB (fairness-to-buyer) and FBS (fairness-between-sellers), both in [0,1]. These signals are blended into the agents' pay-offs through scheduled coefficients $\lambda_{\{buy\}(t)}$ and $\lambda_{\{peer\}(t)}$ that ramp from 0 to 1 over the course of training:

$$R_{\{S_i\}} = r_{\{S_i\}}^{\{raw\}} + \lambda_{\{buy\}(t)} \cdot w_B \cdot \left(\frac{\Sigma_j FTB_j}{N_B}\right) + \lambda_{\{peer\}(t)} \cdot w_P \cdot FBS\left(\frac{\Sigma_j b_i^{\{(j)\}}}{\Sigma_{\{i',j\}} b_{i'}^{\{(j)\}}}\right) \tag{6}$$

$$R_{\{B_j\}} = r_{\{B_j\}}^{\{raw\}} + \lambda_{\{buy\}(t)} \cdot w_B \cdot FTB_j \tag{7}$$

Here, $w_B$ and $w_P$ set the strength of the buyer-fairness and peer-fairness bonuses, respectively, while the peer-fairness bonus is distributed in proportion to *Seller i's* share of the total units sold. This LLM-guided shaping nudges agents toward outcomes that (i) satisfy every buyer's demand, (ii) keep effective prices reasonable, and (iii) balance profits and market share across sellers—complementing the raw-reward structure defined in Equations (4)–(5).

## 2.4 Operational constraints & targets

For every *buyer j*: $\Sigma_i p_i b_i^{\{(j)\}} \leq 7.6 \cdot D_j, \Sigma_i b_i^{\{(j)\}} \leq D_j$, and $b_i^{\{(j)\}} \leq q_i$. For every *seller i*: $q_i \leq I_i$. Training is successful when ≥ 90 % total demand is met, average FTB and FBS ≥ 0.80, seller margins remain in the 20–30 % range, and no seller exceeds 60 % share of total sales.

## 3. Operational Framework: LLM-in-the-Loop MARL

To translate the formal design of Section 2 into a working system, we embed a large-language-model fairness critic directly inside the MARL training loop. Figure 1 sketches the architecture of the decentralized MARL framework, where each episode flows through multiple tightly coupled stages: episode rollout (including sellers posting offers, buyers allocating demand, environment updates, and outcome serialization), deterministic Prompt-Skeleton serialization of episode outcomes, real-time LLM fairness scoring producing FTB and FBS metrics, adaptive reward shaping with validation and error handling, followed by an Independent PPO (IPPO) policy update using shaped rewards. These stages form a closed loop learning cycle that integrates fairness feedback directly into the multi-agent training process.

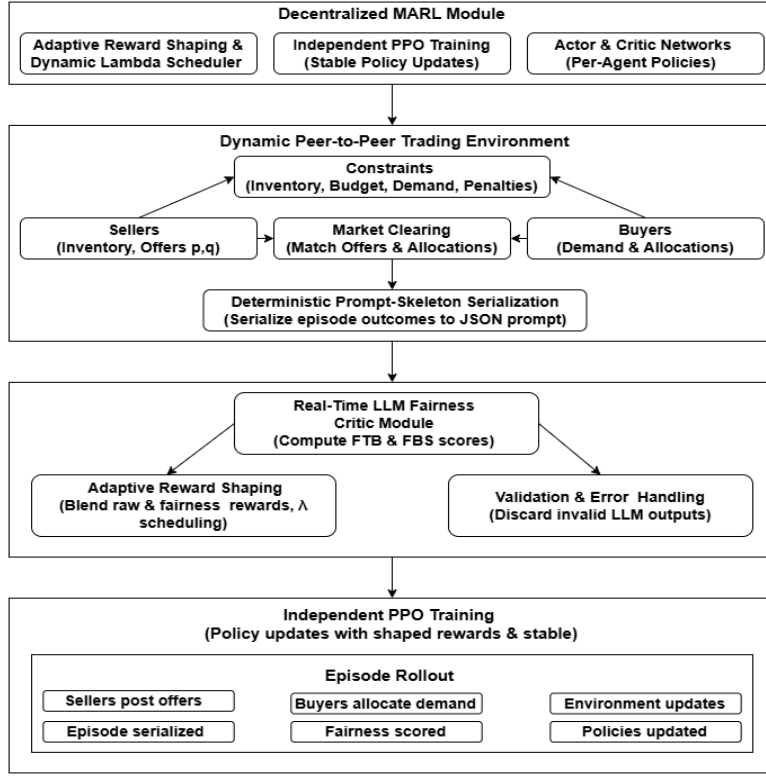

*Figure 1: Architecture of the Decentralized MARL Framework with LLM-Guided Fairness Shaping for Peer-to-Peer Trading*

a) **Episode rollout.** Sellers $S_1 - S_{\{N_S\}}$ sequentially post price quantity offers without knowledge of later sellers' decisions. Buyers $B_1 - B_{\{N\_B\}}$ then allocate their demand across the posted offers. The environment returns each seller's profit, each buyer's expenditure, and any residual demand shortfall following the deterministic dynamics in Section 2.1.

b) **Outcome summarisation & LLM query.** At the end of the episode the entire ledger is serialised into a single, deterministic prompt. An example prompt and the required JSON format are shown below. One LLM call returns $N_B$ buyer-specific FTB scores and one global FBS score; episodes with an invalid LLM response are discarded, ensuring the training signal remains free of heuristic substitutions.

c) **Reward shaping & coefficient schedule.** Shaped rewards are computed exactly as in *Section* 2.3 (*Eqs.* 6–7). $\lambda\_buy(t)$ ramps from 0→1 during the first 20 % of training, while $\lambda\_peer(t)$ ramps from 30 % → 80 %, allowing agents to internalise profitability before full fairness pressure is applied.

d) **IPPO update.** Policies for all sellers and buyers are updated with independent PPO, costing only $O(N_S + N_B)$. $w_B$ and $w_P$ tune the buyer- and peer-fairness bonuses; the latter is split according to each seller's share of units sold. This live LLM signal, without any hand-crafted sharing drives the market to > 90 % demand fulfilment, fair prices, and balanced seller profits, showcasing the first real-time LLM fairness critic in decentralised MARL.

## 4. Simulation Results

To ground the general multi-buyer, multi-seller framework in a concrete setting, we present a controlled case study with the non-trivial topology: two competing sellers and one budget-constrained buyer. Although modest, this configuration retains the core strategic tensions, price competition, inventory allocation, and buyer welfare that motivate our LLM-guided shaping mechanism. All hyper-parameters follow the defaults introduced in Sections 2.3 and 3.

*Table 1: Case-Study Environment and Training Parameters*

| Symbol / Item | Value or Range | Comment |
|---|---|---|
| Seller-1 inventory $I_1$ | 8 – 25 units | Sampled i.i.d. each episode |
| Seller-2 inventory $I_2$ | 10 – 30 units | More stock (same cost) |
| Buyer demand $D$ | 20 – 50 units | Uniform distribution |
| Training horizon | 20,000 episodes | On-policy IPPO |
| Discount factor $\gamma$ | 0.95 | All agents |
| $\lambda_{buy}$, $\lambda_{peer}$ | Linear ramps 0→1 | See Sections 2.3 and 3 |

### 4.1 Learning Dynamics

Figure 2 (left) shows how economic incentives evolve in our system, plotting 500-episode moving-average returns for the two sellers and the buyer, while Figure 3 (right) traces the LLM-supplied fairness signals that modulate those incentives. During the profit-only warm-up ($\lambda = 0$), the sellers exploit their pricing freedom, lifting their own rewards at the buyer's expense. Once buyer-fairness $weight\ \lambda_{buy}$ begins its ascent (marker A, around episode 2,000), both sellers respond by trimming prices and eliminating shortfalls, and the buyer's return rebounds. A second inflection emerges when the peer-fairness weight $\lambda_{peer}$ saturates (marker B, around episode 4,500): the two seller curves collapse onto one another, evidencing the profit equalisation enforced by the peer signal. The fairness trajectories echo this story. Fairness-to-Buyer (FTB) and Fairness-Between-Sellers (FBS) inch upward early on, dip during the transition, and then surge past the 0.80 target, ultimately stabilising above 0.85. Crucially, the turning points in FTB and FBS coincide exactly with the scheduled λ ramps, underscoring real-time LLM feedback rather than static heuristics is what steers the agents toward equitable market outcomes.

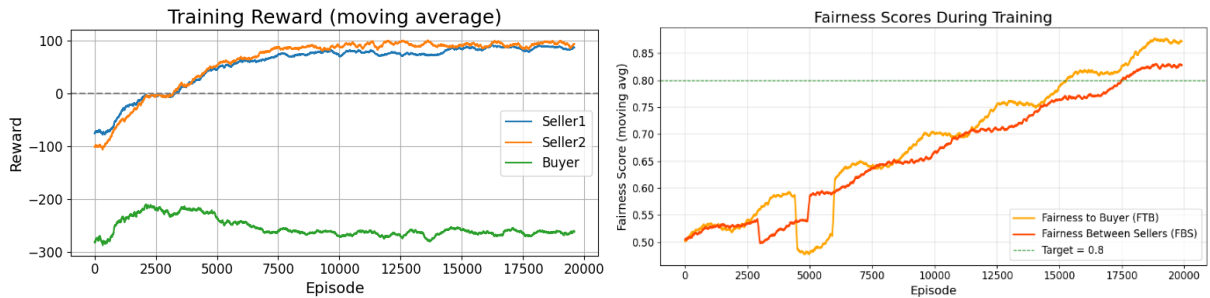

*Figure 2. Learning dynamics for the two-seller, one-buyer case study: Left—agent rewards over training; Right—LLM-derived fairness scores (FTB, FBS).*

### 4.2 Key Performance Indicators

Over the final 2,000 episodes, the system achieves strong performance across all targets. Fairness scores remain high, demand is reliably met, seller profits are balanced, and buyer budgets are respected, demonstrating the stability and effectiveness of our LLM-guided shaping.

*Table 2: Final Evaluation Results*

| Metric | Result | Target |
|---|---|---|
| *Episodes with full demand met* | *92.1 %* | *≥ 90 %* |
| *Average FTB* | *0.88* | *≥ 0.80* |
| *Average FBS* | *0.87* | *≥ 0.80* |
| *Seller margins* | *24 – 26 %* | *20 – 30 %* |
| *Max seller sales share* | *57 %* | *≤ 60 %* |
| *Buyer budget violations* | *0* | *0* |

### 4.3 Ablation: No-LLM Baseline

To assess the contribution of the LLM-based fairness shaping, we perform an ablation in which both shaping coefficients are held at zero $(\lambda_{buy} = 0, \lambda_{peer} = 0)$, effectively disabling the language-model critic. In this setting, fairness metrics plateau at approximately 0.35–0.40, demand fulfilment declines to about 70 %, and *Seller* 2 outperforms *Seller* 1 by roughly 35 % in average profit. These degradations underscore that real-time LLM feedback is not merely beneficial but essential for achieving equitable and efficient outcomes.

### 5. Potential Applications of the Proposed Framework in Power and Energy Systems

The operation of modern power distribution systems has become increasingly complex due to the widespread integration of Distributed Energy Resources (DERs). Unlike traditional unidirectional energy flow paradigms, DERs introduce bidirectional power exchange capabilities, allowing entities to act as prosumers—agents that can both consume and generate electricity. These prosumers can dynamically shift their operational roles by independently buying from or selling power to the utility grid or other entities in the network. As a result, the distribution system evolves into a highly dynamic and decentralized environment, necessitating new frameworks for efficient and fair market regulation.

A key challenge in this context is the absence of structured mechanisms for fair P2P energy trading at the distribution level. Without regulatory support or intelligent coordination, such decentralized interactions may lead to market inefficiencies, unfair pricing dynamics, or operational stress on the Distribution System Operator (DSO). The proposed FairMarket-RL framework addresses this challenge by offering a scalable, fairness-aware, and autonomous trading control solution tailored to multi-agent environments.

An initial and practical application of the framework is in isolated community microgrids (MGs), where individual households can function as independent buying or selling agents. The framework ensures fair transaction outcomes for both buyers and sellers, guided by real-time language-model-based fairness evaluation and reward shaping in reinforcement learning agents. This promotes local energy exchange, enhances system flexibility, and reduces dependency on central grid support, ultimately mitigating operational burdens on DSOs.

Moreover, with the integration of Large Language Models (LLMs) acting as domain-informed fairness critics, the framework exhibits strong scalability. Its architecture can be readily extended to manage large-scale multi-agent trading environments with thousands of prosumers across an interconnected distribution system. Such capability paves the way for practical deployment in future smart grids where autonomous, equitable energy trading is essential for both economic efficiency and grid stability.

### 6.Conclusion

This paper introduces FairMarket-RL, a language-model-guided reward-shaping framework for multi-agent reinforcement learning in peer-to-peer markets. By embedding an instruction-tuned Large Language Model as a real-time fairness critic, this approach replaces brittle, hand-crafted shaping rules with dense, human-interpretable feedback. In a controlled two-seller/one-buyer case study, the resulting Independent PPO agents reliably fulfilled more than 90% of demand, achieved average fairness scores (FTB about 0.88 and FBS about 0.87), and prevented profit monopolization, all while outperforming a no-LLM baseline.

Conceptually, FairMarket-RL demonstrates that natural-language fairness assessments can be injected directly into the training loop of non-textual environments, effectively bridging the gap between normative objectives and continuous control. Practically, the framework scales to many buyers and sellers, making it suitable for emerging applications such as DER-rich microgrids, gig-economy platforms, and digital asset exchanges. Building on these promising results, future work includes on device distillation of the LLM critic for low-latency deployment, expanding the fairness vocabulary to cover sustainability and grid-stability objectives, and conducting formal robustness analyses against adversarial or strategically manipulated prompts. Together, these efforts aim to bring socially aligned autonomous trading systems closer to real-world decentralized markets.